\documentclass[a4paper,conference]{IEEEtran}
\IEEEoverridecommandlockouts

\usepackage{cite}
\usepackage{amsmath,amssymb,amsfonts}
\usepackage{algorithmic}
\usepackage{graphicx}
\usepackage{textcomp}
\usepackage[table,xcdraw]{xcolor}
\usepackage{booktabs} 
\usepackage{xcolor}
\usepackage{booktabs}
\usepackage{colortbl}
\usepackage{multirow}
\usepackage{graphicx}
\usepackage{url}
\usepackage[normalem]{ulem}
\usepackage{hyperref}
\usepackage{authblk}
\usepackage{cleveref}

\def\BibTeX{{\rm B\kern-.05em{\sc i\kern-.025em b}\kern-.08em
    T\kern-.1667em\lower.7ex\hbox{E}\kern-.125emX}}
\begin{document}

\title{Self-Adaptive Gamma Context-Aware SSM-based Model for Metal Defect Detection}

\author{
\textbf{Sijin Sun}$^{1,5,\dagger}$ \hspace{10pt}
\textbf{Ming Deng}$^{2,\dagger}$ \hspace{10pt}
\textbf{Xingrui Yu}$^{1,3}$ \hspace{10pt}
\textbf{Xingyu Xi}$^4$ \hspace{10pt}
\textbf{Liangbin Zhao}$^{1,*}$
}
\affil{\normalsize
$^\dagger$Equal contribution \quad $^*$Corresponding author \\
$^1$ Institute of High Performance Computing, Agency for Science, Technology and Research (A*STAR IHPC), Singapore \\
$^2$ Shanghai University \\ 
$^3$ Centre for Frontier AI Research, Agency for Science, Technology and Research (A*STAR CFAR), Singapore \\
$^4$ Shanghai Maritime University \quad $^5$
National University of Singapore
}

\maketitle

\begin{abstract}
Metal defect detection is critical in industrial quality assurance, yet existing methods struggle with grayscale variations and complex defect states, limiting its robustness. To address these challenges, this paper proposes a Self-Adaptive Gamma Context-Aware SSM-based model(GCM-DET). Proposed detection framework integrating a Dynamic Gamma Correction (GC) module to enhance grayscale representation and optimize feature extraction for precise defect reconstruction. A State-Space Search Management (SSM) architecture captures robust multi-scale features, effectively handling defects of varying shapes and scales. Focal Loss is employed to mitigate class imbalance and refine detection accuracy. Additionally, the CD5-DET dataset is introduced, specifically designed for port container maintenance, featuring significant grayscale variations and intricate defect patterns. Experimental results demonstrate that the proposed model achieves substantial improvements, with mAP@0.5 gains of 27.6\%, 6.6\%, and 2.6\% on the CD5-DET, NEU-DET, and GC10-DET datasets.
Dataset available at \url{https://universe.roboflow.com/stanleysun233/cd5-det/dataset/1}.


\end{abstract}


\section{Introduction}
%
The quality of metal surfaces is critical in various industrial applications, including aerospace, manufacturing, and container transportation. Surface defects, such as cracks, dents, and scratches, not only compromise the structural integrity and aesthetics of metal products but also lead to significant economic losses if left undetected. As a result, the accurate and efficient detection of metal surface defects has become an essential task in industrial quality control.

In recent years, the adoption of deep learning techniques has significantly advanced the performance of defect detection systems\cite{terven2023comprehensive}. Convolutional neural networks (CNNs) and transformer-based models have demonstrated exceptional capabilities in handling complex image-based tasks, enabling automated and reliable defect detection. However, several challenges remain: 1) Metal defect often exhibits varied and localized features, making effective multi-scale feature aggregation vital for improving detection accuracy. 2) The boundary features of damaged areas are typically irregular and complex. 3) Extreme recognition bias due to environment and perspective.

To address these challenges, this paper proposes novel improvements in two key modules of the defect detection pipeline, enhancing both the precision and robustness of detection methods. Inspired by the CARAFE\cite{wang2019carafe}, a gamma coefficient downstream correction module based on dynamic coefficients is proposed. Besides, state space model\cite{zhu2024vision} reconstructs the weights of the feature map through content perception to obtain the deep features of the model. Then, the F-Loss\cite{zhang2022focal} is introduced to weigh the indicators between single weights and global weights. 

Furthermore, CD5-DET is designed for container damage detection, containing a diverse set of labeled images that capture 5 defect types under varying environmental conditions. The dataset bridges a critical gap in the field by providing a standardized benchmark for defect detection in industrial settings.

\begin{figure}
    \centering
    \includegraphics[width=1.0\linewidth]{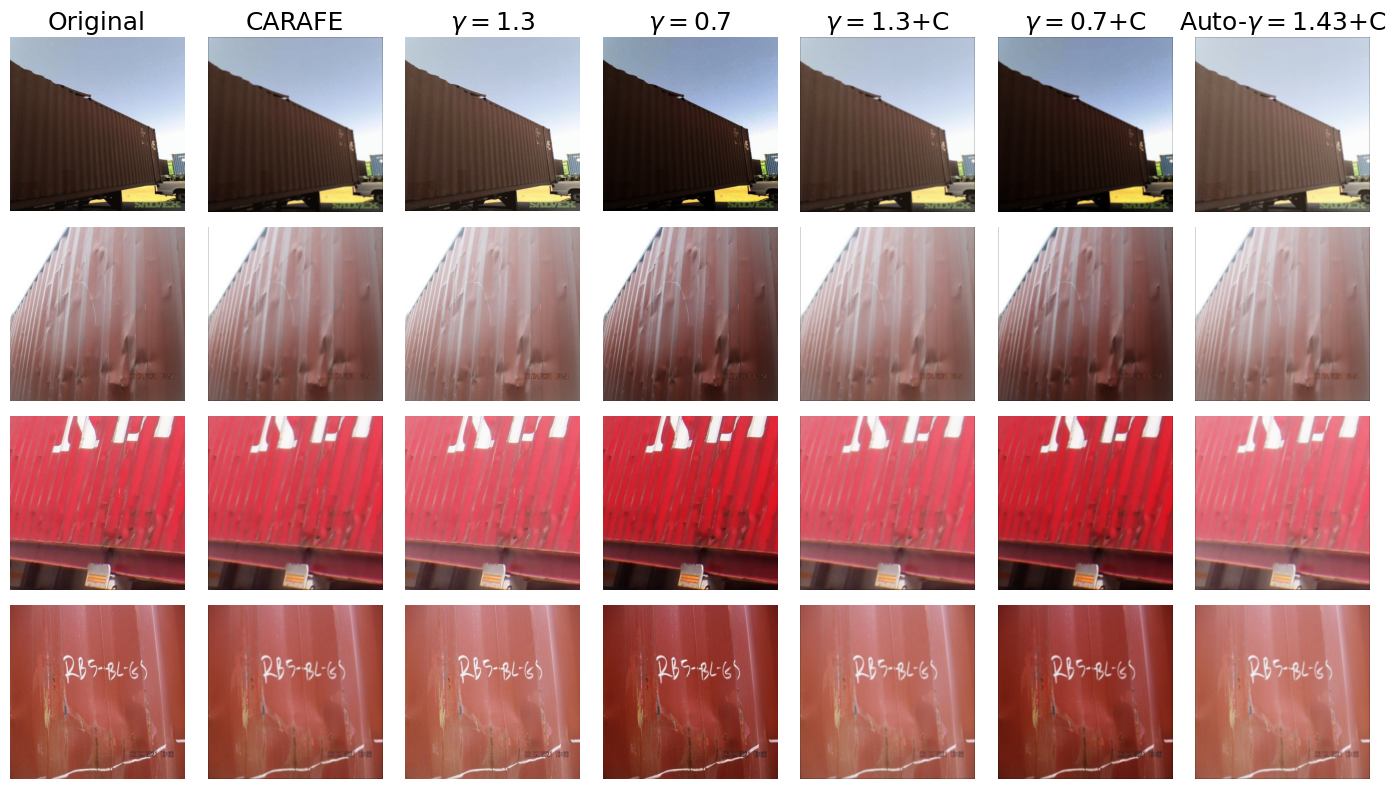}
    \caption{Using different Gamma $\gamma$ coefficients can enhance image quality by adjusting exposure and shadows. Multiplying pixel values by $\gamma$ applies a linear scaling transformation. A Gamma coefficient greater than 1 brightens the image, while one less than 1 darkens it. This adjustment improves the visibility of the image in specific situations.}
    \label{fig:show-difference}
\end{figure}

\begin{figure*}[!htbp]
    \centering
    \includegraphics[width=1.0\linewidth]{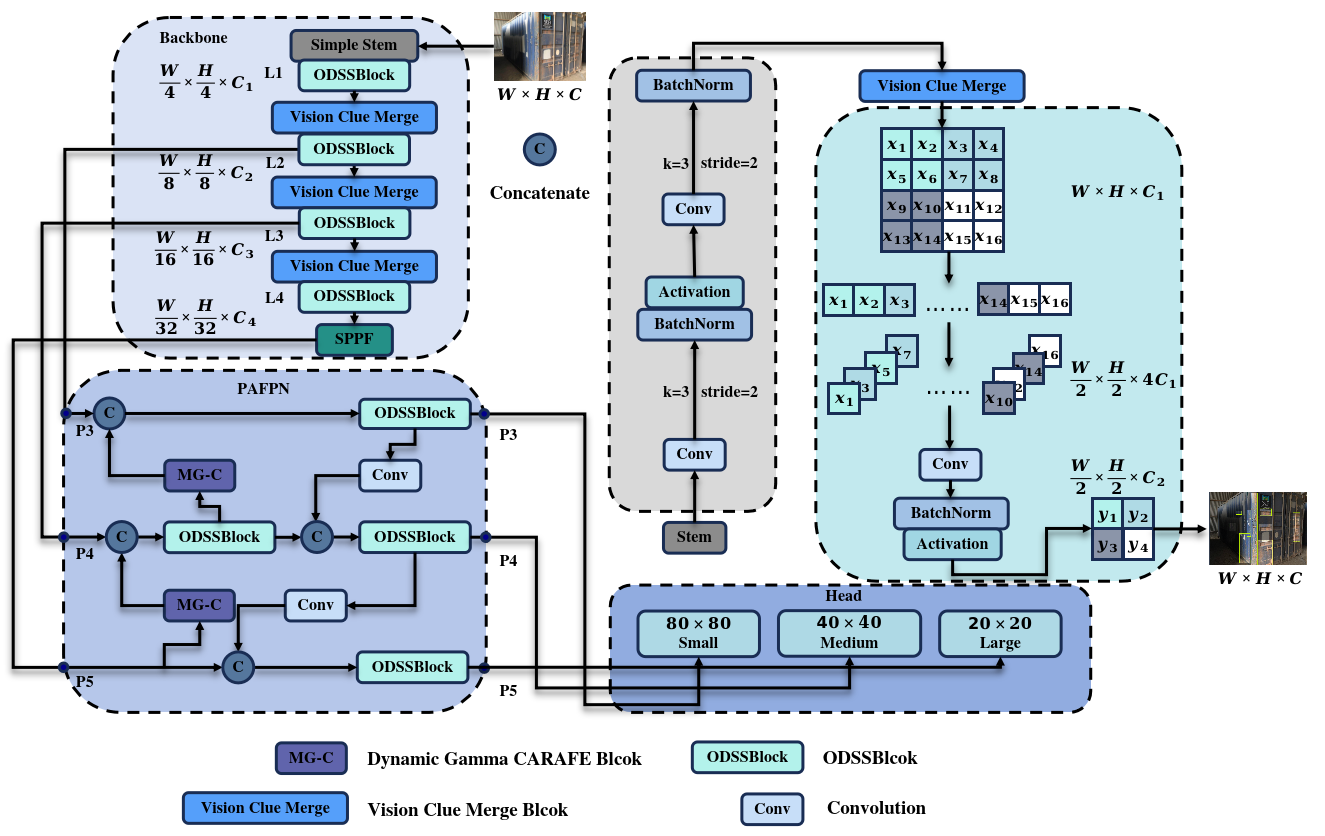}
    \caption{Proposed framework includes a Dynamic Gamma Block, where the Gam-Car model applies adaptive Gamma correction to the original image based on variations in $\mu$ and $\sigma$, while the Pix-shuffling provides super-resolution sampling for image scaling. An ODSS Block and a Vision Clue Merge Block are employed to further enhance the image processing.}
    \label{fig:yolomambacarafe}
\end{figure*}

The main contributions of this study can be summarized as follows.
\begin{enumerate}
\renewcommand\labelenumi{(\theenumi)}
    \item This paper proposes a upsampling structure combined with a dynamic gamma coefficient.
    
    \item This paper proposes a GCM-DET model based on the State-space search architecture, which is used to improve the backbone and neck of the model.
    
    \item This work collected and developed the industry's first container damage dataset CD5-DET, and then obtained excellent indicators on the well-known defect detection datasets NEU-DET and GC10-DET.
\end{enumerate}

\section{Related Work}
\label{sec:related-work}
Several metal defect detection datasets, such as NEU-DET \cite{song2013noise} and GC10-DET \cite{lv2020deep}, have been widely used in research. Common baseline models include the YOLO series, Faster-RCNN, and U-Net. Typically, an engineering computer with a camera captures images, which are then processed and input into detection algorithms \cite{chen2024efficient}. Similarly, in container defect detection, models are designed, evaluated, and deployed in port camera systems. Thus, developing efficient algorithms and high-quality datasets is crucial for target detection. 

Early defect detection relied on machine learning models using handcrafted features. Wang \cite{wang2009hog} combined HOG-LBP for optimal detection on the INRIA dataset. Song et al. \cite{song2013noise} improved noise processing with an optimized LBP and SVM classifier. Li et al. \cite{li2013learning} replaced the Haar classifier \cite{wilson2006facial} with SURF for target detection. These methods achieved high accuracy in simple tasks using manual feature extraction and statistical priors.

Advancements in computer vision have enhanced deep learning for target detection, categorized into two-step and single-step methods. Two-step approaches, such as RCNN \cite{girshick2014rich}, Fast-RCNN \cite{ren2015faster}, and Mask-RCNN \cite{he2017mask}, first generate candidate boxes before classification. Single-step methods, like the YOLO series \cite{hussain2024yolo,sun2024st}, detect and classify objects directly using CNNs. Related research also considers the detection based on LLM of prompt\cite{xiong2025enhancing}. In addition, processing detection tasks from image frequency features\cite{li2025fedkd,deng2025fmnet} is also a direction. Deep learning enables automatic feature extraction, outperforming traditional defect detection in complex environments with higher accuracy and speed.

Due to YOLO’s strong feature capture capabilities, many scholars have proposed improvements. Woo et al. \cite{woo2018cbam} introduced the CBAM module for attention inference, while SE attention \cite{wu2023yolo} and ViTransformer \cite{zhang2021vit} enhanced detection accuracy and robustness. ShuffleNet \cite{zhang2018shufflenet} improved efficiency with pointwise group convolution, and MobileNet \cite{howard2017mobilenets} balanced model size and performance using width and resolution multipliers. The Ghost module \cite{zhang2023real} accelerated YOLO’s backbone computation.

Proposed work compares with the above models under identical training settings to highlight the superior multi-scale representation which this method achieves.

\section{Methodology}
\label{sec:methodology}
\subsection{Detection Framework}

YOLOv8 models, especially YOLOv8n, are used for target detection due to their balance of accuracy and speed \cite{jocher2023yolo}. The Visual State Space Model \cite{zhu2024vision,zeng2025enhancing} reduces computational complexity, while the Gamma-Carafe structure replaces the upsampling layer for noisy, low-resolution images. Focal-IoU loss addresses dynamic factor differences. The resulting detection framework is shown in \Cref{fig:yolomambacarafe}.

\subsection{Gamma-Carafe}
Inspired by the CARAFE (Content-Aware ReAssembly of FEatures) structure, this work introduces dynamic grayscale correction to replace the original interpolation upsampling layer of the baseline model. CARAFE is a content-aware upsampling method that reorganizes input features by generating dynamic convolution kernels, making the reconstructed high-resolution feature map more accurate. In addition, dynamic Gamma processing optimizes the robustness input of the model.

Improved Gam-Car model is added a dynamic search gamma method before implementing CARAFE. In order to reduce the model's dependence on the grayscale of the image, dynamically adjusting the model's brightness coefficient can increase the model's ability to capture deep features.

The implementation of Gamma-Carafe mainly consists of three steps: 1) Optimize the grayscale features of the model; 2) Generate predicted reassembly kernels; 3) Reassemble kernel features.

\paragraph{Dynamic Gamma Correction} This algorithm adjusts image brightness dynamically by employing a Gamma correction that is computed based on the image's brightness characteristics. First, the input RGB image tensor $ x $ is converted to grayscale to derive brightness information. This grayscale brightness is calculated using the weighted sum \Cref{eq:gray}: 

\begin{equation}
    gray = 0.299 \cdot R + 0.587 \cdot G + 0.114 \cdot B 
    \label{eq:gray}
\end{equation}
, where $ R $, $ G $, and $ B $ represent the red, green, and blue channels, respectively.

To adaptively determine the Gamma value, the mean and standard deviation of the brightness are calculated for each image in the batch. These values are used to modulate Gamma dynamically: 

\begin{equation}
    \gamma = G_{min} +  (G_{max} - G_{min}) \cdot  \frac{gray_{mean}}{ gray_{mean}+ gray_{std} + \epsilon}
\label{eq:gamr}
\end{equation}

where $\epsilon$ is a small constant to prevent division by zero. The resulting Gamma value is clamped to ensure it remains within the pre-defined range $[G_{min}, G_{max}]$.

Finally, the calculated Gamma is applied to each pixel of the image by raising the pixel values (incremented by a small value $\epsilon$) to the power of Gamma, normalizing the brightness without exceeding the valid range $[0, 1]$. This dynamic approach allows the algorithm to adjust the image brightness more effectively, reducing reliance on grayscale characteristics alone and enhancing feature retention. The effects of different $\gamma$ can be seen in \Cref{fig:show-difference}.

\paragraph{Kernel Prediction.}
For kernel prediction, the purpose of this module is to predict the appropriate convolution kernel $W_{l'}$. By predicting the adjacent region $N(X_l, k_{\text{encoder}})$ in position $l'$, it can be calculated:
\begin{equation}
\mathcal W_{l'}=\psi(N(\mathcal X_l,k_{encoder}))
\label{eq:carafe-kp}
\end{equation}

The \Cref{eq:carafe-kp} generates an adaptive convolution kernel based on the input feature content, which is used to dynamically reorganize the features. Here $N(X_l, k_{\text{encoder}})$ represents the information of the input feature $X_l$ in the neighborhood of the convolution kernel size $k_{\text{encoder}}$.
\paragraph{Feature Reassembly.}

Use $\psi$ to infer the adaptive convolution kernel $W_{l'}$, and resample the features through $N(X_l, k_{\text{up}})$, expressed as:
\begin{equation}
X'_{l'} = \phi(N(X_l, k_{\text{up}}), W_{l'})
\label{eq:carafe-ra}
\end{equation}

The \Cref{eq:carafe-ra} uses the content-aware convolution kernel to reorganize input features, so that the upsampling process can be adaptively adjusted according to the content at different locations, thus improving upsampling accuracy and feature expression ability. The steps of feature reorganization are as follows.

\paragraph{Channel Compression.} 
The module performs channel compression on the input features, and compresses the number of input channels $c$ to the number of intermediate channels $c_{\text{mid}}$ through convolution operation:
\begin{equation}
F_{\text{comp}}(x) = W_{\text{comp}} * x + b_{\text{comp}}
\label{eq:carafe-fc}
\end{equation}
Where $W_{\text{comp}}$ is the convolution kernel used for compression, $b_{\text{comp}}$ is the bias term, and * indicates the convolution operation. This operation compresses the number of feature channels from $c$ to $c_{\text{mid}}$.
\paragraph{Content Encoder.} 
Next, a convolution layer is used to generate the upsampling kernel. The size of the generated convolution kernel is $(scale \times k_{\text{up}})^2$, where $scale$ is the upsampling factor and $k_{\text{up}}$ is the kernel size used for upsampling. This process is expressed by the following equation:
\begin{equation}
F_{\text{enc}}(x) = W_{\text{enc}} * F_{\text{comp}}(x)
\label{eq:carafe-kg}
\end{equation}
$W_{\text{enc}}$ is the weight used to generate the convolution kernel. The dimension of the output $F_{\text{enc}}(x)$ is $(scale \times k_{\text{up}})^2$.
\paragraph{Pixel Normalization.} 
After the generated convolution kernel, the pixel shuffling operation is used to rearrange the generated feature map into a high-resolution map. Assume that the input feature map is $F_{\text{in}} \in \mathbb{R}^{H \times W \times C}$, where $C = C_{\text{out}} \times r^2$. The \texttt{PixelShuffle} process can be expressed as:
\begin{equation}
\resizebox{\columnwidth}{!}{$
F_{\text{out}}(i, j, k) = F_{\text{in}}\left(\left\lfloor \frac{i}{r} \right\rfloor, \left\lfloor \frac{j}{r} \right\rfloor, C_{\text{out}} \times (i \mod r) + (j \mod r) + k \right)$}
\label{eq:carafe-ps}
\end{equation}
The PixelShuffle operation re-arranges the input channels so that the feature map resolution is increased by the upsampling factor.

Finally, combining convolution kernel perception and feature reorganization by \Cref{eq:carafe-tt}.
\begin{equation}
\mathcal{X}_{l^{\prime}}^{\prime}=\sum_{n=-r}^r\sum_{m=-r}^r\mathcal{W}_{l^{\prime}(n,m)}\cdot\mathcal{X}_{(i+n,j+m)}
\label{eq:carafe-tt}
\end{equation}

\subsection{State Space Model}
The structured state space models (S4)\cite{Gu2021EfficientlyML} and Mamba\cite{Gu2023MambaLS}, as instances of State Space Model (SSM), are derived from continuous systems. These models map a one-dimensional input function or sequence $x(t) \in \mathbb{R}$ to an output $y(t) \in \mathbb{R}$ thought a hidden state $h(t) \in \mathbb{R}^N$. The evolution parameter $A \in \mathbb{R}^{N \times N}$ and  projections parameters $B \in \mathbb{R}^{N \times 1}$ , $C \in \mathbb{R}^{1 \times N}$ are used in this system.

\begin{equation}
h'(t) = \mathbf{A} h(t) + \mathbf{B} x(t) 
\label{eq:state_update}
\end{equation}
\begin{equation}
y(t) = \mathbf{C} h(t)
\label{eq:outy}
\end{equation}

S4 and Mamba are discrete versions of the continuous system. They use a timescale parameter $\Delta$ to discretize the continuous parameters $A$ and $B$ into $\bar{A}$ and $\bar{B}$. The commonly used method for this transformation is zero-order hold (ZOH), which is defined as follows:
\begin{equation}
\bar{\mathbf{A}} = \exp(\Delta \mathbf{A})
\end{equation}

\begin{equation}
\bar{\mathbf{B}} = (\Delta \mathbf{A})^{-1} \left( \exp(\Delta \mathbf{A}) - \mathbf{I} \right) \Delta \mathbf{B}
\end{equation}

After discretizing the parameters into $\bar{A}$ and $\bar{B}$ with a step size of $\Delta$, the \Cref{eq:state_update} and \Cref{eq:outy} can be rewritten as:

\begin{equation}
h_t = \bar{\mathbf{A}} h_{t-1} + \bar{\mathbf{B}} x_t
\end{equation}

\begin{equation}
y_t = \mathbf{C} h_t 
\end{equation}

Finally, the model computes the output through a global convolution.
\begin{equation}
\bar{\mathbf{K}} = (\mathbf{C} \bar{\mathbf{B}}, \mathbf{C} \bar{\mathbf{A}} \bar{\mathbf{B}}, \ldots, \mathbf{C} \bar{\mathbf{A}}^{L-1} \bar{\mathbf{B}})
\end{equation}

\begin{equation}
y = x * \bar{\mathbf{K}}
\end{equation}

Where $\bar{K} \in \mathbb{R}^L$ is the structured convolutional  kernel, and $L$ is the length of the input sequence $x$.

\begin{figure}[!htbp]
    \centering
    \includegraphics[width=\linewidth]{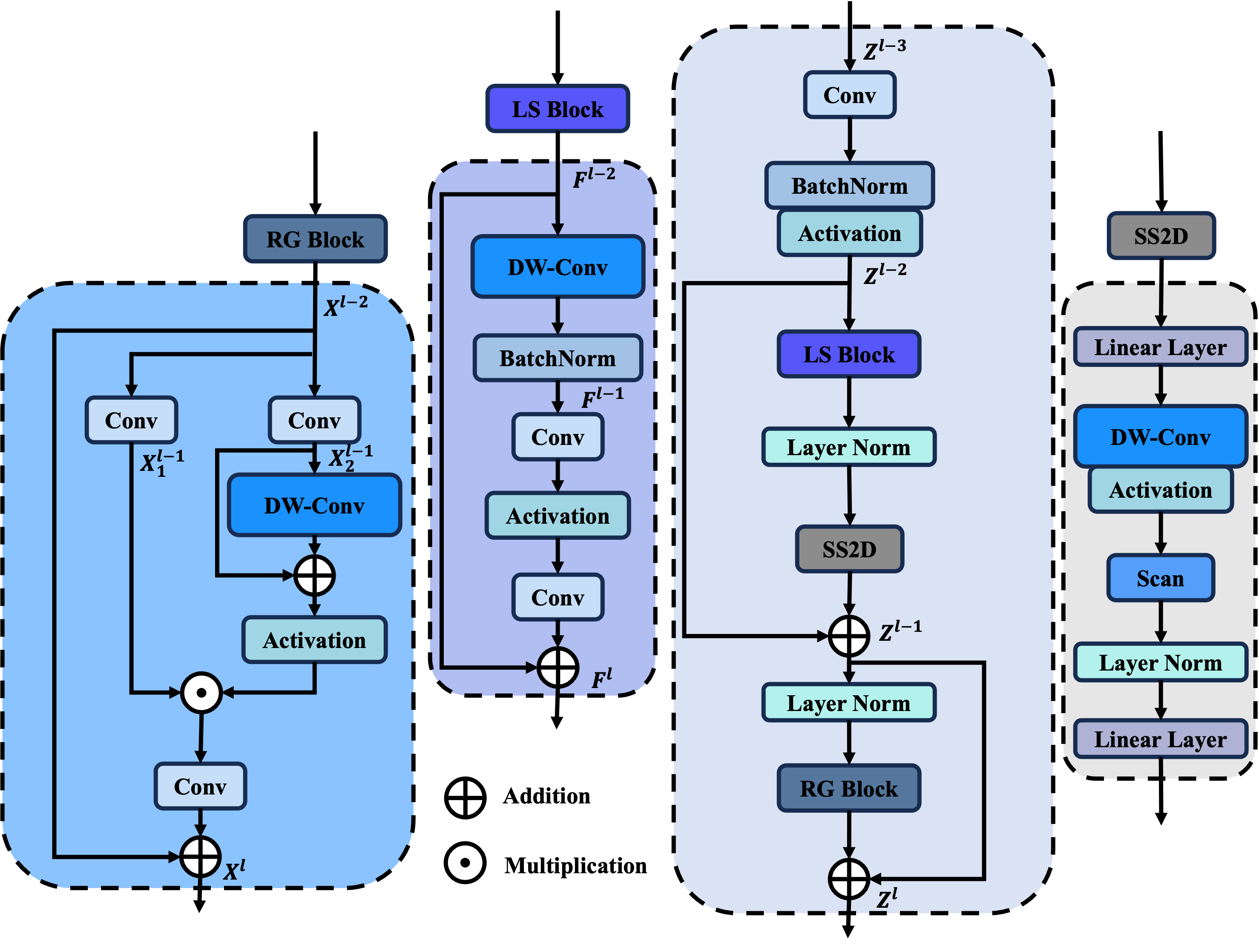}
    \caption{ODSSBlock}
    \label{fig:ODSSBlock}
\end{figure}

Mamba-YOLO\cite{Wang2024MambaYS} can be referred to as \Cref{fig:yolomambacarafe}, Mamba-YOLO replaces the backbone with ODMamba, which consists of a Simple Stem using two convolutions with a stride of 2 and a kernel size of 3, along with a downsampling module. The neck follows the PAN-FPN design, using the ODSSBlock module to replace C2f\cite{Chen2023YOLOMSRM}, where Conv is solely responsible for downsampling. The backbone first performs downsampling through the Stem module.The Stem module uses two convolution operations with a stride of 2 and a kernel size of 3, generating a 2D feature map with a resolution of $H/4 \times W/4$, followed by further downsampling through the Vision Clue Merge and ODSSBlock modules. The Vision Clue Merge module employs a method of splitting feature maps and using 1x1 convolutions to reduce dimensions while retaining more visual clues, thereby optimizing the training of State Space Models (SSMs). By removing normalization, splitting the dimension maps, and utilizing 4x compressed pointwise convolutions, it simplifies the processing flow while preserving more critical feature information, enhancing the model’s performance and efficiency. ODSSBlock as shown in the \Cref{fig:ODSSBlock}.

ODSSBlock performs a series of processes in the input stage, while maintaining efficiency and stability in the training and inference process through batch normalization.

Compared with standard convolution or attention mechanisms, the state-space formulation can model long-range spatial interactions and subtle local features simultaneously. By discretizing the continuous state evolution with a suitable $\Delta$, the SSM captures spatiotemporal dependencies inherent in textures and edges. This is particularly beneficial for complex defect shapes where boundary irregularities demand a more global yet flexible representation. For instance, when detecting random corrosion spots, the structured convolution kernel derived by SSM can adapt to changing patterns without inflating computational costs, thereby retaining both global context and minute details.

\begin{equation}
Z^{l-2} = \hat{\Phi}(BN(Conv_{1\times1}(Z^{l-3})))
\end{equation}

Where $\hat{\Phi}$ represents the activation function (nonlinear SiLU), inspired by the architecture style of Transformer Blocks \cite{Dosovitskiy2020AnII}, the ODSSBlock incorporates layer normalization and residual connections. The equations are as follows:

\begin{equation}
Z^{l-1} = SS2D(LN(LS(Z^{l-2}))) + Z^{l-2}
\end{equation}

\begin{equation}
Z^{l} = RG(LN(Z^{l-1})) + Z^{l-1}
\end{equation}

Where LS and RG represent LocalSpatial Block and ResGated Block, respectively, $Z^{l-3}$ and $Z^{l}$ denote the input and output features, and $Z^{l-1}$ represents the intermediate state after 2D-Selective-Scan (SS2D \cite{Liu2024VMambaVS}).

LocalSpatial Block is used to enhance the capture of local features. First, depth-separable convolution is applied to the given input feature $F^{l-1} \in \mathbb{R}^{C \times H \times W}$, operating independently on each input channel without mixing the channel information to effectively extract the local spatial information of the input feature map, while simultaneously reducing computational cost and the number of parameters. Then, Batch Normalization is performed to provide regularization and avoid overfitting. The resulting intermediate state $F^{l-1}$ is defined as:

\begin{equation}
F^{l-1} = BN(DWConv_{3 \times 3}(F^{l-2}))
\end{equation}

The intermediate state $F^{l-1}$ mixes the channel information through a 1×1 convolution, using the nonlinear GELU activation function to better preserve the information distribution. 

\begin{equation}
F^l = Conv_{1 \times 1}(\Phi(Conv_{1 \times 1}(F^{l-1}))) \oplus F^{l-2}
\end{equation}

Where $F^l$ is the output feature, and $\Phi$ represents the activation function.

The RG Block creates two branches, $X_1^{l-1}$ and $X_2^{l-1}$, from the input $X^{l-2}$, and each branch implements a fully connected layer in the form of a 1x1 convolution.

\begin{equation}
X_1^{l-1} = \text{Conv}_{1 \times 1}(X^{l-2})  
\end{equation}

\begin{equation}
X_2^{l-1} = \text{Conv}_{1 \times 1}(X^{l-2})
\end{equation}

The RG Block adopts a nonlinear GeLU as the activation function, followed by elemental multiplication to fuse with the $X_2^{l-1}$ branch, and a $1 \times 1$ convolution to extract global features and blend channel information. Finally, the features in the hidden layer are added to the original input $X^{l-2}$ through residual connections. With only a slight increase in computational cost, the RG Block gains the ability to capture more global features, and the feature $X^l$ is defined as follows:

\begin{multline}
X^l = \text{Conv}_{1 \times 1}(X_1^{l-1} \odot  \Phi(\text{DWConv}_{3 \times 3}(X_2^{l-1}) \\
\oplus X_2^{l-1}))   
\oplus X^{l-2}
\end{multline}

Where $\Phi$ denotes the activation function. The gating mechanism of the RG Block uses integrated convolution operations, preserving the spatial information in the image and making the model more sensitive to fine-grained features.

\section{Experiment}
\label{sec:experiment}

\subsection{Datasets Preparation}
In this study, for container defect detection, the CD5-DET\cite{sun2024cd5det} is used, which contains deframe, hole, minior\_dent, major\_dent and rust. In addition, NEU-DET\cite{song2013noise} and GC10-DET\cite{lv2020deep} are collected for generalization experiments and ablation experiments.

\subsection{Experimental Environment}
A host machine is operated with Docker virtualization for the experiment with RTX-4090. The batch size is set to 16, the initial learning rate is set to 0.01, and the training cycle is set to 300. All input image data are uniformly processed to an input image size of 640 × 640. The dataset is divided according to the default ratio published by the corresponding author.

\begin{table*}[!htbp]
\centering
\caption{Comparison Study on CD5-DET}
\begin{tabular}{@{}cccccccccccc@{}}
\toprule
\multicolumn{1}{c}{\multirow{2}{*}{Method}} & \multicolumn{5}{c}{AP@0.5} & \multicolumn{1}{c}{\multirow{2}{*}{mAP@0.5}} & \multicolumn{1}{c}{\multirow{2}{*}{mAP@0.5:0.95}} & \multicolumn{1}{c}{\multirow{2}{*}{Params(M)}} & \multicolumn{1}{c}{\multirow{2}{*}{GFLOPs}} & \multicolumn{1}{c}{\multirow{2}{*}{Size(Mb)}} \\
\cmidrule{2-6}
                        & \multicolumn{1}{c}{De.} & \multicolumn{1}{c}{Ho.} & \multicolumn{1}{c}{Ma.} & \multicolumn{1}{c}{Mi.} & \multicolumn{1}{c}{Ru.} &                          &                               &                            &                         &                           \\ 
\midrule
\multicolumn{1}{c|}{YOLOv5n}                 & 0.257                & 0.589          & 0.199                & 0.054          & 0.155         & 0.251                    & \multicolumn{1}{c|}{0.119}                         & 2.5                        & 7.1                     & 5.0                       \\
\multicolumn{1}{c|}{YOLOv6n}                 & 0.177                & 0.552          & 0.133                & 0.006          & 0.073         & 0.188                    & \multicolumn{1}{c|}{0.082}                         & 4.2                        & 11.8                    & 8.3                       \\
\multicolumn{1}{c|}{YOLOv8n}                 & 0.336                & 0.548          & 0.183                & 0.069          & 0.151         & 0.257                    & \multicolumn{1}{c|}{0.132}                         & 3.0                        & 8.1                     & 6.0                       \\
\multicolumn{1}{c|}{YOLOv8s}                 & \textbf{0.503}       & \textbf{0.656} & \underline{0.281}    & \underline{0.151} & \underline{0.283} & \textbf{0.375}   & \multicolumn{1}{c|}{\textbf{0.203}}                & 11.1                       & 28.4                    & 21.4                      \\
\multicolumn{1}{c|}{YOLOv8l}                 & 0.428                & \underline{0.594} & 0.247             & \textbf{0.190} & \textbf{0.339} & \underline{0.360}      & \multicolumn{1}{c|}{\underline{0.191}}             & 43.6                       & 165.4                   & 166.9                     \\
\multicolumn{1}{c|}{YOLOv8-world}            & \underline{0.434}    & 0.589          & 0.226                & 0.121          & 0.281         & 0.330                    & \multicolumn{1}{c|}{0.171}                         & 4.1                        & 14.2                    & 15.8                      \\
\multicolumn{1}{c|}{YOLOv8-worldv2}          & 0.406                & 0.482          & 0.241                & 0.082          & 0.234         & 0.289                    & \multicolumn{1}{c|}{0.134}                         & 3.5                        & 9.9                     & 13.9                      \\
\multicolumn{1}{c|}{YOLOv8-ghost}            & 0.404                & 0.549          & 0.256                & 0.054          & 0.146         & 0.282                    & \multicolumn{1}{c|}{0.138}                         & 1.7                        & 5.1                     & 3.6                       \\
\multicolumn{1}{c|}{YOLOv10n}                & 0.128                & 0.500          & 0.125                & 0.039          & 0.092         & 0.177                    & \multicolumn{1}{c|}{0.081}                         & 2.7                        & 8.4                     & 10.8                      \\
\multicolumn{1}{c|}{YOLOv10s}                & 0.259                & 0.548          & \textbf{0.288}       & 0.101          & 0.196         & 0.278                    & \multicolumn{1}{c|}{0.177}                         & 8.1                        & 24.8                    & 31.3                      \\ 
\midrule
\multicolumn{1}{c|}{YOLO-Mamba-B}            & \underline{0.600}    & 0.568          & \underline{0.382}    & \underline{0.316} & \underline{0.376} & \textbf{0.448}    & \multicolumn{1}{c|}{\textbf{0.246}}                & 21.8                       & 49.7                    & 41.9                      \\
\multicolumn{1}{c|}{YOLO-Mamba-T}            & 0.415                & \textbf{0.595} & 0.361                & 0.106          & 0.274         & 0.350                    & \multicolumn{1}{c|}{0.173}                         & 6.1                        & 14.3                    & 11.7                      \\
\multicolumn{1}{c|}{YOLO-Mamba-L}            & 0.436                & 0.542          & 0.325                & 0.245          & 0.372         & \underline{0.485}        & \multicolumn{1}{c|}{\underline{0.284}}             & 57.6                       & 156.2                   & 110.5                     \\ 
\midrule
\multicolumn{1}{c|}{GCM-DET(Ours)}           & \textbf{0.660}       & \underline{0.590} & \textbf{0.430}    & \textbf{0.386} & \textbf{0.596} & \textbf{0.533}         & \multicolumn{1}{c|}{\textbf{0.374}}                & 21.8                       & 49.7                    & 83.9                      \\ 
\bottomrule
\end{tabular}
\label{tab:comp-cd5-det}
\end{table*}

\subsection{Evaluation Metrics}

The study evaluates model performance based on accuracy and complexity. Accuracy is measured using AP@T and mAP@T, with common thresholds at 50\% and 50:95\%. mAP@0.5 considers predictions correct when the predicted box $g^p$ overlaps the ground truth $g^t$ by at least 50\%, while mAP@50:95 provides a stricter accuracy assessment (\Cref{eq:ap}, \Cref{eq:map}). Complexity is assessed using parameter count (spatial complexity), GFLOPs (computational cost), and model size (storage requirements).

\begin{equation}
    AP\left(i\right)=\int_{0}^{1}P\left(R\right)\mathrm{d}R
    \label{eq:ap}
\end{equation}

\begin{equation}
    mAP=\frac{\sum_{i=1}^{n}AP(i)}{n}
    \label{eq:map}
\end{equation}


Focal IoU Loss in \Cref{eq:fiou-fiou} is used to evaluate the training loss:
\begin{equation}
\mathcal{L}_{\text{Focal IoU}} = \sum_{i=1}^{N} \alpha_i \cdot (1 - \text{IoU}(B_{p,i}, B_{g,i}))^\gamma
\label{eq:fiou-fiou}
\end{equation}
Where \( N \) represents the number of samples, \( \alpha_i \) is the weight of the \( i \)th sample, and \( \text{IoU}(B_{p,i}, B_{g,i}) \) represents the IoU of the \( i \)th sample.

\subsection{Comparison Study}

GCM-DET outperforms multiple YOLO baselines and the World CLIP method, achieving the highest AP@0.5 in De. (0.660), Ma. (0.430), Mi. (0.386), and Ru. (0.596), while ranking second in Ho. to Mamba-T. It leads in overall mAP@0.5 (0.533) and mAP@0.5:0.95 (0.374). Among YOLO models, YOLOv8s performs best. GCM-DET halves computation compared to YOLOv8l and improves accuracy over Mamba-L while reducing computation. It achieves this without adding much more parameters compared to Mamba-B.

\begin{figure}[!htbp]
    \centering
    \includegraphics[width=\linewidth]{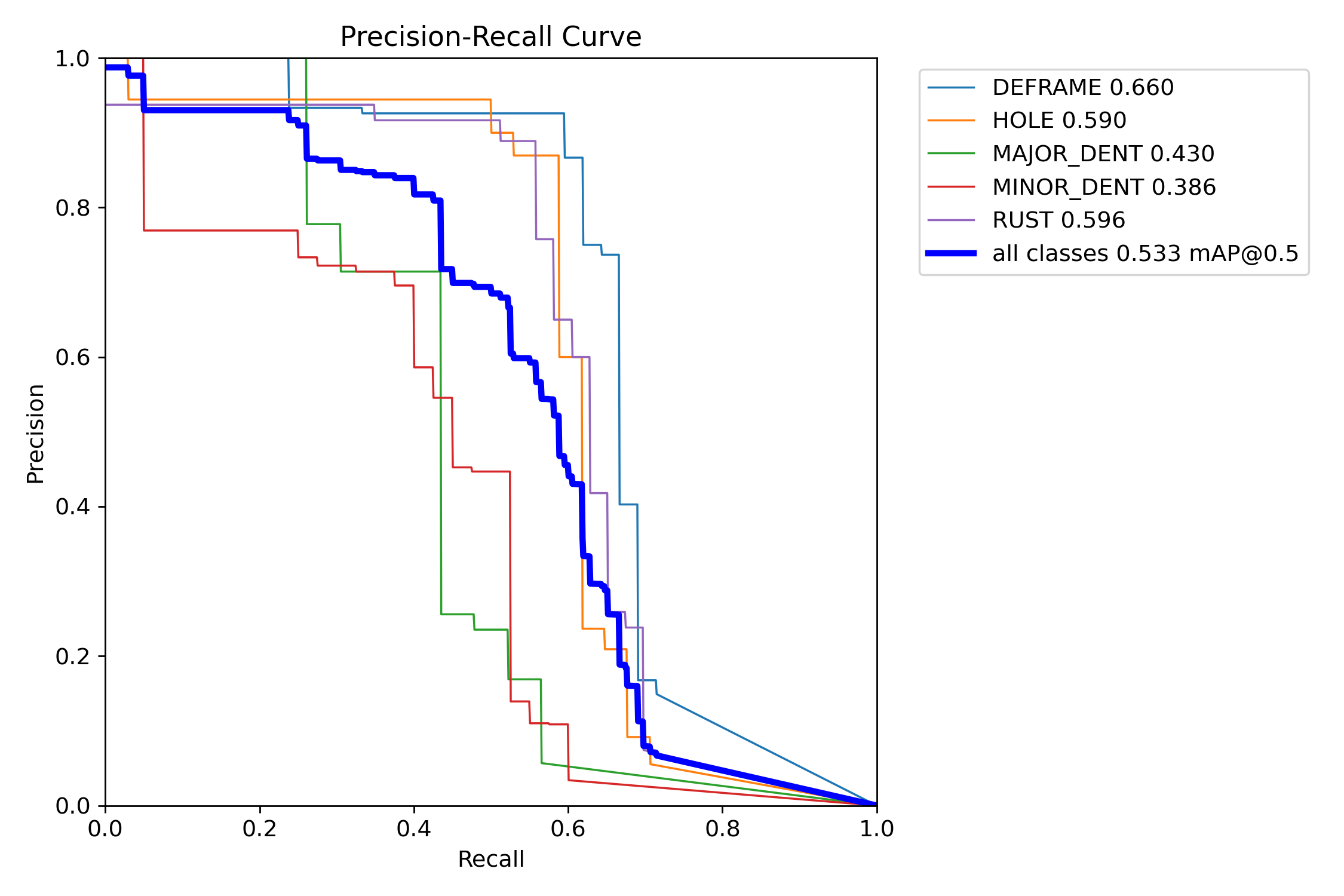}
    \caption{Precision-recall metrics for the GCM-DET model.}
    \label{fig:gcm-prc}
\end{figure}

Based on the \Cref{fig:gcm-prc}, it can be observed that there are significant differences in the detection performance of different defect categories: the average precision (mAP@0.5) of DEFRAME, RUST and HOLE is relatively high, indicating that the model has strong robustness in the location and classification of these types of defects; while MAJOR\_DENT and MINOR\_DENT are relatively weak, and the accuracy of the curve in the medium and high recall range is significantly reduced, indicating that it is more susceptible to noise interference or feature confusion.

\begin{table*}[!htbp]
\centering
\caption{Detection Results of Different Models in NEU-DET}
\begin{tabular}{@{}cccccccccccc@{}}
\toprule
\multirow{2}{*}{Method}              & \multicolumn{6}{c}{AP@0.5}                                                                          & \multirow{2}{*}{mAP@0.5}            & \multirow{2}{*}{mAP@0.5:0.95} & \multirow{2}{*}{Params(M)} & \multirow{2}{*}{GFLOPs} & \multirow{2}{*}{Size(Mb)} \\ \cmidrule(lr){2-7}
 & Cr.            & In.            & Pa.            & Ps.            & Rs.            & Sc.            &                                     &                               &                            &                         &                           \\ \midrule
\multicolumn{1}{c|}{RCNN-ResNet50}   & 0.462          & 0.838          & 0.895          & 0.875          & 0.558          & 0.905          & \multicolumn{1}{c}{0.756}          & \multicolumn{1}{c|}{-}                             & 41.37                      & 13.40                   & -                         \\
\multicolumn{1}{c|}{RCNN-ResNet101}  & 0.586          & 0.875          & 0.879          & 0.895          & 0.603          & 0.936          & \multicolumn{1}{c}{0.796}          & \multicolumn{1}{c|}{-}                             & 60.37                      & 18.20                   & -                         \\ \midrule
\multicolumn{1}{c|}{DF-DERT}         & 0.440          & 0.814          & 0.933          & 0.875          & 0.701          & 0.947          & \multicolumn{1}{c}{0.785}          & \multicolumn{1}{c|}{-}                             & 41.00                      & 13.60                   & -                         \\
\multicolumn{1}{c|}{Focus-DERT}      & 0.553          & 0.849          & 0.942          & 0.826          & 0.634          & 0.957          & \multicolumn{1}{c}{0.793}          & \multicolumn{1}{c|}{-}                             & 49.20                      & 11.30                   & -                         \\ \midrule
\multicolumn{1}{c|}{ES-Net}          & 0.560          & \textbf{0.876} & 0.883          & 0.874          & 0.604          & 0.949          & \multicolumn{1}{c}{0.791}          & \multicolumn{1}{c|}{-}                             & 147.98                     & -                       & -                         \\
\multicolumn{1}{c|}{DEA}             & 0.609          & 0.825          & 0.943          & 0.958          & 0.672          & 0.741          & \multicolumn{1}{c}{0.791}          & \multicolumn{1}{c|}{-}                             & 42.20                      & -                       & -                         \\ \midrule
\multicolumn{1}{c|}{YOLOv5n}         & 0.494          & 0.852          & 0.920          & 0.830          & 0.607          & 0.907          & \multicolumn{1}{c}{0.768}          & \multicolumn{1}{c|}{0.456}                         & 2.50                       & 7.10                    & 5.03                      \\
\multicolumn{1}{c|}{YOLOv5s}         & 0.527          & 0.806          & 0.888          & 0.861          & 0.616          & 0.939          & \multicolumn{1}{c}{0.773}          & \multicolumn{1}{c|}{0.452}                         & 9.11                       & 23.80                   & 17.60                     \\
\multicolumn{1}{c|}{YOLOv6n}         & 0.437          & 0.846          & 0.926          & 0.847          & 0.631          & 0.925          & \multicolumn{1}{c}{0.769}          & \multicolumn{1}{c|}{0.459}                         & 4.23                       & 11.80                   & 8.30                      \\
\multicolumn{1}{c|}{YOLOv6s}         & 0.507          & 0.852          & 0.888          & 0.830          & 0.653          & 0.896          & \multicolumn{1}{c}{0.771}          & \multicolumn{1}{c|}{0.439}                         & 16.30                      & 44.00                   & 31.30                     \\
\multicolumn{1}{c|}{YOLOv7n-tiny}    & 0.451          & 0.798          & 0.904          & 0.869          & 0.616          & 0.827          & \multicolumn{1}{c}{0.744}          & \multicolumn{1}{c|}{0.379}                         & 6.02                       & 13.10                   & 11.70                     \\
\multicolumn{1}{c|}{YOLOv8n}         & 0.473          & 0.843          & 0.923          & 0.823          & 0.624          & 0.926          & \multicolumn{1}{c}{0.769}          & \multicolumn{1}{c|}{0.444}                         & 3.01                       & 8.10                    & 5.97                      \\
\multicolumn{1}{c|}{YOLOv8s}         & 0.508          & 0.840          & 0.908          & 0.847          & 0.619          & 0.926          & \multicolumn{1}{c}{0.775}          & \multicolumn{1}{c|}{0.454}                         & 11.10                      & 28.40                   & 21.40                     \\ \midrule
\multicolumn{1}{c|}{LiFSO-Net}       & 0.481          & 0.864          & 0.942          & 0.880          & 0.644          & 0.942          & \multicolumn{1}{c}{0.792}          & \multicolumn{1}{c|}{0.479}                         & \textbf{1.77}                       & \textbf{5.70}                    & \textbf{3.69}                      \\
\multicolumn{1}{c|}{RDD-YOLO}        & 0.529          & 0.859          & 0.944          & 0.862          & 0.707          & 0.966          & \multicolumn{1}{c}{0.811}          & \multicolumn{1}{c|}{-}                             & -                 & -              & -                \\
\multicolumn{1}{c|}{MD-YOLO}         & 0.467          & 0.814          & 0.913          & 0.851          & \textbf{0.726} & 0.920          & \multicolumn{1}{c}{0.782}          & \multicolumn{1}{c|}{-}                             & 9.00                       & 14.10                   & -                \\ \midrule
\multicolumn{1}{c|}{GCM-DET(Ours)} & \textbf{0.619} & 0.727          & \textbf{0.980} & \textbf{0.996} & 0.536          & \textbf{0.979} & \multicolumn{1}{c}{\textbf{0.835}} & \multicolumn{1}{c|}{\textbf{0.496}}                & 21.80                      & 49.70                   & 83.90                     \\ \bottomrule
\end{tabular}
\label{table:neu-acc}
\end{table*}



\subsection{Generalization Study}

GCM-DET is evaluated on the NEU-DET dataset with various detection models in \Cref{table:neu-acc}, including YOLOs, U-Net, RCNN, and improved models like LiFSO-Net and RDD-YOLO. It outperforms others in mAP@0.5 (0.835), surpassing RDD-YOLO (0.811) and RCNN-ResNet101 (0.796). Compared to lightweight models like YOLOv5n (0.768) and YOLOv8n (0.769), GCM-DET improves accuracy by 6.7\% and 6.6\%. For mAP@0.5:0.95, it achieves 0.473, slightly behind LiFSO-Net (0.479) but ahead of YOLOv8n (0.444), demonstrating strong generalization. It excels in category-wise performance, leading in 4 of 6 categories and achieving near-perfect detection (98\%+) in Pa., Ps., and Sc., proving its robustness.

\subsection{Ablation Study}
In the ablation experiments, baseline model YOLOv8n is compared marked as S.N.1 in \Cref{table:ab-cd5},\ref{table:ab-neu},\ref{table:ab-gc10}. And its different improved versions (M-DET, G-DET, GC-DET and GCM-DET) in three datasets (CD5-DET, NEU-DET and GC10- DET).



\begin{table}[!htbp]
\centering
\caption{Ablation Experiments Performed on CD5-DET.}
\resizebox{\columnwidth}{!}{
\begin{tabular}{cccccccc}
\toprule
\multicolumn{1}{c}{\multirow{2}{*}{S.N.}} & \multicolumn{3}{c}{Method} & \multicolumn{4}{c}{Metric} \\
\cmidrule(r){2-4} \cmidrule(l){5-8}
& \multicolumn{1}{c}{Gamma} & \multicolumn{1}{c}{Carafe} & \multicolumn{1}{c}{SSM} & \multicolumn{1}{c}{mAP@0.5} & \multicolumn{1}{c}{mAP@0.5:0.95} & \multicolumn{1}{c}{Precision} & \multicolumn{1}{c}{Recall} \\ 
\midrule
\multicolumn{1}{c|}{1} & & & & \multicolumn{1}{|c}{0.257} & \multicolumn{1}{c}{0.132} & \multicolumn{1}{c}{0.378} & \multicolumn{1}{c}{0.288} \\
\multicolumn{1}{c|}{2} & & & $\surd$ & \multicolumn{1}{|c}{0.447} & \multicolumn{1}{c}{0.243} & \multicolumn{1}{c}{0.622} & \multicolumn{1}{c}{0.458} \\
\multicolumn{1}{c|}{3} & $\surd$ & & & \multicolumn{1}{|c}{0.463} & \multicolumn{1}{c}{0.237} & \multicolumn{1}{c}{0.624} & \multicolumn{1}{c}{0.503} \\
\multicolumn{1}{c|}{4} & $\surd$ & $\surd$ & & \multicolumn{1}{|c}{0.509} & \multicolumn{1}{c}{0.318} & \multicolumn{1}{c}{0.709} & \multicolumn{1}{c}{0.510} \\ 
\midrule
\multicolumn{1}{c|}{\textbf{Ours}} & $\surd$ & $\surd$ & $\surd$ & \multicolumn{1}{|c}{\textbf{0.533}} & \multicolumn{1}{c}{\textbf{0.374}} & \multicolumn{1}{c}{\textbf{0.522}} & \multicolumn{1}{c}{\textbf{0.754}} \\ 
\bottomrule
\end{tabular}
}
\label{table:ab-cd5}
\end{table}


\begin{table}[!htbp]
\centering
\caption{Ablation Experiments Performed on NEU-DET.}
\resizebox{\columnwidth}{!}{
\begin{tabular}{cccccccc}
\toprule
\multicolumn{1}{c}{\multirow{2}{*}{S.N.}} & \multicolumn{3}{c}{Method} & \multicolumn{4}{c}{Metric} \\
\cmidrule(r){2-4} \cmidrule(l){5-8}
& \multicolumn{1}{c}{Gamma} & \multicolumn{1}{c}{Carafe} & \multicolumn{1}{c}{SSM} & \multicolumn{1}{c}{mAP@0.5} & \multicolumn{1}{c}{mAP@0.5:0.95} & \multicolumn{1}{c}{Precision} & \multicolumn{1}{c}{Recall} \\ 
\midrule
\multicolumn{1}{c|}{1} & & & & \multicolumn{1}{|c}{0.769} & \multicolumn{1}{c}{0.444} & \multicolumn{1}{c}{0.666} & \multicolumn{1}{c}{0.720} \\
\multicolumn{1}{c|}{2} & & & $\surd$ & \multicolumn{1}{|c}{0.817} & \multicolumn{1}{c}{0.465} & \multicolumn{1}{c}{\textbf{0.771}} & \multicolumn{1}{c}{0.754} \\
\multicolumn{1}{c|}{3} & $\surd$ & & & \multicolumn{1}{|c}{0.784} & \multicolumn{1}{c}{0.449} & \multicolumn{1}{c}{0.673} & \multicolumn{1}{c}{\textbf{0.785}} \\
\multicolumn{1}{c|}{4} & $\surd$ & $\surd$ & & \multicolumn{1}{|c}{0.806} & \multicolumn{1}{c}{0.438} & \multicolumn{1}{c}{0.734} & \multicolumn{1}{c}{0.762} \\ 
\midrule
\multicolumn{1}{c|}{\textbf{Ours}} & $\surd$ & $\surd$ & $\surd$ & \multicolumn{1}{|c}{\textbf{0.835}} & \multicolumn{1}{c}{\textbf{0.473}} & \multicolumn{1}{c}{0.751} & \multicolumn{1}{c}{0.763} \\ 
\bottomrule
\end{tabular}
}
\label{table:ab-neu}
\end{table}




\begin{table}[!htbp]
\centering
\caption{Ablation Experiments Performed on GC10-DET.}
\resizebox{\columnwidth}{!}{
\begin{tabular}{cccccccc}
\toprule
\multicolumn{1}{c}{\multirow{2}{*}{S.N.}} & \multicolumn{3}{c}{Method} & \multicolumn{4}{c}{Metric} \\
\cmidrule(r){2-4} \cmidrule(l){5-8}
& \multicolumn{1}{c}{Gamma} & \multicolumn{1}{c}{Carafe} & \multicolumn{1}{c}{SSM} & \multicolumn{1}{c}{mAP@0.5} & \multicolumn{1}{c}{mAP@0.5:0.95} & \multicolumn{1}{c}{Precision} & \multicolumn{1}{c}{Recall} \\ 
\midrule
\multicolumn{1}{c|}{1} & & & & \multicolumn{1}{|c}{0.639} & \multicolumn{1}{c}{0.320} & \multicolumn{1}{c}{0.632} & \multicolumn{1}{c}{0.634} \\
\multicolumn{1}{c|}{2} & & & $\surd$ & \multicolumn{1}{|c}{0.657} & \multicolumn{1}{c}{0.320} & \multicolumn{1}{c}{0.657} & \multicolumn{1}{c}{0.626} \\
\multicolumn{1}{c|}{3} & $\surd$ & & & \multicolumn{1}{|c}{0.646} & \multicolumn{1}{c}{\textbf{0.326}} & \multicolumn{1}{c}{0.677} & \multicolumn{1}{c}{0.624} \\
\multicolumn{1}{c|}{4} & $\surd$ & $\surd$ & & \multicolumn{1}{|c}{0.645} & \multicolumn{1}{c}{0.325} & \multicolumn{1}{c}{0.670} & \multicolumn{1}{c}{0.612} \\ 
\midrule
\multicolumn{1}{c|}{\textbf{Ours}} & $\surd$ & $\surd$ & $\surd$ & \multicolumn{1}{|c}{\textbf{0.665}} & \multicolumn{1}{c}{\textbf{0.326}} & \multicolumn{1}{c}{\textbf{0.681}} & \multicolumn{1}{c}{\textbf{0.636}} \\ 
\bottomrule
\end{tabular}
}
\label{table:ab-gc10}
\end{table}

1) \textbf{Remove the SSM module and verify whether the state-space search method is effective.} Comparing the baseline model (S.N.1) and SSM (S.N.2) models in \Cref{table:ab-cd5}, \Cref{table:ab-neu} and \Cref{table:ab-gc10}, SSM achieved a 19.0\% performance improvement in the CD5-DET dataset, a 4.8\% performance improvement in NEU-DET, and a 1.8\% improvement in GC10-DET on the mAP@0.5 indicator.

2) \textbf{Remove the dynamic gamma module and verify whether the gamma coefficient correction improves the indicators.} The baseline model and the Gamma (S.N.3) model have improved performance in all three datasets, indicating that the dynamically changing Gamma coefficient can repair the grayscale features of the model. Then, the experiment shows that the Carafe (S.N.3) model can obtain more upsampling features compared with the Gamma-Carafe (S.N.4) module with grayscale correction added.

3) \textbf{Verify whether the fusion model is effective.} By merging the backbone network of the SSM architecture and the Gamma-Carafe coefficient, proposed model achieved all-round performance improvement.

\subsection{Sensitivity Analysis of SSMs}

\Cref{tab:size-mamba} shows the specific parameters of Mamba models.

\begin{table}[!htbp]
\centering
\caption{Window size for different parameters.}
\begin{tabular}{cccc}
\toprule
\multicolumn{1}{c}{\multirow{2}{*}{model}} & \multicolumn{3}{c}{parameter} \\
\cmidrule(l){2-4}
& \multicolumn{1}{c}{depth} & \multicolumn{1}{c}{width} & \multicolumn{1}{c}{max channels} \\ 
\midrule
\multicolumn{1}{c|}{YOLO-Mamba-T} & 0.33 & 0.25 & 1024 \\
\multicolumn{1}{c|}{YOLO-Mamba-B} & 0.33 & 0.5 & 1024 \\
\multicolumn{1}{c|}{YOLO-Mamba-L} & 0.67 & 0.75 & 768 \\ 
\bottomrule
\end{tabular}
\label{tab:size-mamba}
\end{table}

\begin{figure}
    \centering
    \includegraphics[width=1.0\linewidth]{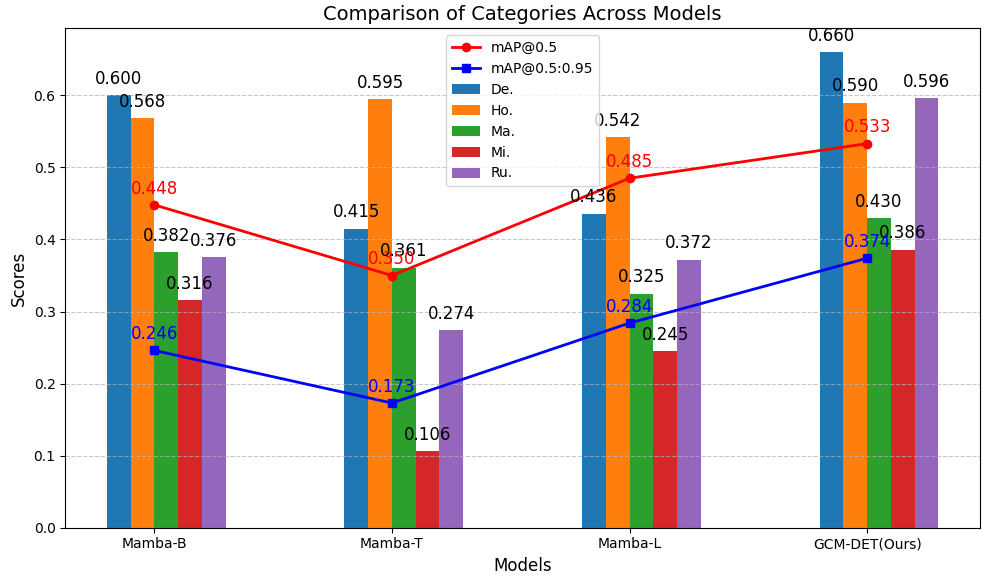}
    \caption{The impact of different window sizes on SSMs performance. Proposed model acheves best in each classes.}
    \label{fig:ssm-parameter}
\end{figure}

The results in the \Cref{fig:ssm-parameter} show that Mamba models of different sizes have different sensitivities to different features. For the container damage detection task, Mamba-B is the best window size.

\subsection{Discussion}

Through comparison of comparative experiments, generalization experiments and ablation experiments, the following are the findings and discussions achieved:

1) GCM-DET shows significant advantages in detection performance, especially in the mAP@0.5 and mAP@0.5:0.95. This performance improvement is mainly attributed to the introduction of multi-scale feature fusion with adaptive gamma correction and the SSM-based mechanism.

2) Compared with more complex models such as MD-YOLO and RDD-YOLO, GCM-DET performs more balanced on the NEU-DET dataset. The combination of adaptive Gamma correction and multi-scale feature fusion enables the model to better adapt to diverse detection tasks.

3) GCM-DET achieves high performance but with greater complexity. Despite the increased computational demand, GCM-DET is designed for tasks requiring high accuracy, improving system reliability and security. Higher detection precision reduces errors in downstream tasks, especially in automated decision-making, enhancing overall system performance.

\section{Conclusion}
\label{sec:conclusion}

GCM-DET improves detection accuracy with an innovative module design and effective feature fusion, excelling in complex scenes and multi-scale target tasks. It achieves a top mAP@0.5 of 0.533 on the CD5-DET dataset, demonstrating strong deep feature capture and enhanced grayscale information. Despite high computational complexity, it offers a reliable solution for high-accuracy tasks. Future research will focus on optimizing efficiency and adaptability to reduce computational costs, while expanding the dataset to include more target types and scenarios, particularly for container defect detection. This will support a more versatile and efficient detection system.

\section*{Acknowledgments}
This research was funded by Programme MARS (Programme of Maritime AI Research in Singapore) with funding grant number SMI-2022-MTP-06 by Singapore Maritime Institute (SMI).


\end{document}